\documentclass[sigplan,twocolumn,nonacm]{acmart}
\acmSubmissionID{<640>}
\renewcommand\footnotetextcopyrightpermission[1]{}
\usepackage{booktabs}
\usepackage{tikz}
\usepackage{comment}
\usepackage{amsmath}
\usepackage{caption}
\usepackage{subcaption}
\usepackage{graphicx}
\usepackage{float}
\usepackage{algorithm}
\usepackage{algorithmicx}
\usepackage[noend]{algpseudocode}
\algnewcommand{\LineComment}[1]{\State \(//\) #1}
\usepackage[normalem]{ulem}
\usepackage{bbding}
\usepackage{todonotes}
\usepackage{filecontents}
\usepackage{minted}
\usepackage{enumitem}
\usepackage[bottom]{footmisc}
\usepackage[]{hyperref}
\usepackage{cleveref}
\usepackage{pifont}
\usepackage{multirow}
\usepackage{xcolor}
\usepackage{outlines}
\usepackage{diagbox}
\usepackage{wasysym}

\definecolor{LightGray}{gray}{0.9}
\definecolor{FadedBanana}{RGB}{255,255,191}
\definecolor{DeepChalk}{RGB}{255,191,191}
\definecolor{FadedFlora}{RGB}{191,255,191}
\definecolor{DeepSnow}{RGB}{191,255,255}
\definecolor{SoapStone}{RGB}{218,218,218}
\definecolor{LightCayenneSixty}{RGB}{239,206,211}
\definecolor{LightCayenne}{RGB}{208,143,145}

\newcommand{\nonsense}[1]{\textcolor{black}{#1}}

\newcommand{\sysname}{{\textsc{Auras}}}

\begin{document}

% \title[]{\sysname{}: Co-locating LLM inference with LLM finetune to Improve GPU utilization}
\title[]{Boosting Embodied AI Agents through Perception-Generation Disaggregation and Asynchronous Pipeline Execution}

\author{Shulai Zhang}
\affiliation{\institution{Shanghai Jiao Tong University}
\city{}\country{}}
% \email{zslzsl1998@sjtu.edu.cn}

\author{Ao Xu}
\affiliation{\institution{Shanghai Jiao Tong University}
\city{}\country{}}
% \email{xuao123@sjtu.edu.cn}

\author{Quan Chen}
\affiliation{\institution{Shanghai Jiao Tong University}
\city{}\country{}}
% \email{chen-quan@cs.sjtu.edu.cn}

\author{Han Zhao}
\affiliation{\institution{Shanghai Jiao Tong University}
\city{}\country{}}
% \email{zhaohan_miven@sjtu.edu.cn}

\author{Weihao Cui}
\affiliation{\institution{Shanghai Jiao Tong University}
\city{}\country{}}
% \email{weihao@sjtu.edu.cn}

\author{Ningxin Zheng}
\affiliation{\institution{Bytedance}
\city{}\country{}}
% \email{zhengningxin@bytedance.com}

\author{Haibin Lin}
\affiliation{\institution{Bytedance}
\city{}\country{}}

\author{Xin Liu}
\affiliation{\institution{Bytedance}
\city{}\country{}}

\author{Minyi Guo}
\affiliation{\institution{Shanghai Jiao Tong University}
\city{}\country{}}
% \email{guo-my@cs.sjtu.edu.cn}

\begin{abstract}
Embodied AI systems operate in dynamic environments, requiring seamless integration of perception and generation modules to process high-frequency input and output demands. Traditional sequential computation patterns, while effective in ensuring accuracy, face significant limitations in achieving the necessary "thinking" frequency for real-world applications. In this work, we present \sysname{}, an algorithm-system co-designed inference framework to optimize the inference frequency of embodied AI agents.
\sysname{} disaggregates the perception and generation and provides controlled pipeline parallelism for them to achieve high and stable throughput.
Faced with the data staleness problem that appears when the parallelism is increased, \sysname{} establishes a public context for perception and generation to share, thereby promising the accuracy of embodied agents.
Experimental results show that \sysname{} improves throughput by $2.54\times$ on average while achieving $102.7\%$ of the original accuracy, demonstrating its efficacy in overcoming the constraints of sequential computation and providing high throughput.
\end{abstract}

\maketitle

\section{Introduction}

Embodied Artificial Intelligence (AI) represents a rapidly evolving field at the intersection of intelligent manufacturing, autonomous driving, and robotics. Unlike LLM-based chatbots~\cite{google2023bard, openai2022chatgpt, chen2021evaluatinglargelanguagemodels, devlin-etal-2019-bert} that operate solely in virtual environments, embodied AI agents interact with the physical world, integrating perception, decision-making, and action in real-time~\cite{rt2, rth2024arxiv, kim24openvla, chi2023diffusion, tinyvla}. 
% These agents are designed to perform tasks that require understanding and responding to dynamic environments, making them a key enabler for applications requiring autonomy and adaptability.
The ability to ``think'' and act in real-time is a fundamental requirement for these applications.
% , demanding innovative solutions at both the algorithmic and system levels.
%Recent advancements in embodied AI have revealed a pivotal observation: 
The cutting-edge capabilities of embodied AI are predominantly driven by generative AI algorithms. Generative models, such as LLMs~\cite{openai2023gpt4, touvron2023llamaopenefficientfoundation, meta-llama2023llama3, schmid2023finetune} and diffusion-based algorithms~\cite{DL_nonequilibriumthermodynamics, chi2023diffusion, Denoisingdiffusionprobabilisticmodels, pmlr-v139-nichol21a}, have emerged as critical components for their superior reasoning abilities. 
% These algorithms excel at tasks requiring complex pattern recognition, contextual understanding, and decision-making, making them particularly well-suited for the intricate demands of embodied AI applications. 
A defining characteristic of generative AI algorithms is their iterative nature~\cite{Attentionisallyouneed, Denoisingdiffusionprobabilisticmodels}, where multiple rounds of generation are employed to achieve high-quality outputs. 
% This iterative process enables these models to produce nuanced and contextually relevant results, but it also imposes significant computational demands.

\begin{figure}
    \centering
    \includegraphics[width=\linewidth]{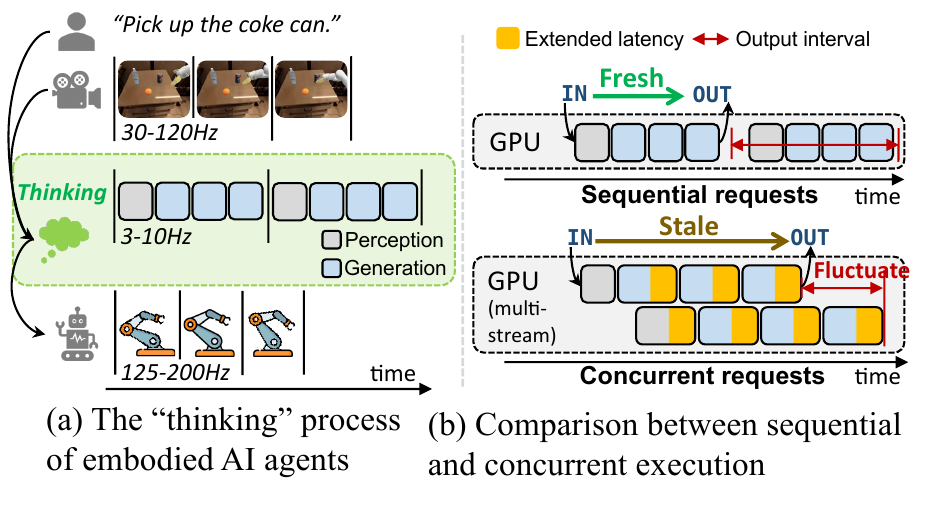}
    \vspace{-6mm}
    \caption{The general workflow of an embodied AI agent.}
    \label{fig:intro}
\end{figure}

\autoref{fig:intro}(a) illustrates the general workflow of current embodied AI systems. These systems process human language prompts and continuous images as multi-modal inputs, utilize generative algorithms for ``thinking'', and output actions for embodied backends, such as robotic actuators. %Additionally, these 
Specifically, each image captured by the camera triggers a request within the embodied system. This request undergoes the thinking process to generate an action for the robotics. Any additional requests generated during this period are discarded until the current request completes the overall action generation process.
Such systems are often deployed on edge devices that typically feature a consumer-level GPU (e.g., an Nvidia RTX 4090) to meet low-latency response requirements. %Such edge devices typically feature a single consumer-level GPU, such as an Nvidia RTX 4090.

With limited resources and advanced algorithms, the ``thinking'' stage suffers from the low-frequency problem. Specifically, input data streams, such as videos captured by high-resolution cameras, often operate at frequencies of 30 to 120 frames per second or higher~\cite{chi2023diffusion, Sanket2019EVDodgeEA, EdgeFlowNet, Araujo_2023_CVPR}. Robotic actuators require output commands at equally high or higher frequencies (e.g., 125-200Hz~\cite{chi2023diffusion}) to achieve smooth and precise movements. 
However, the ``thinking'' frequency of current embodied AI agents is quite low (e.g., 3-10Hz for Google-series embodied agents~\cite{o2023open, rt2} and 1-6Hz for OpenVLA~\cite{kim24openvla}).
% However, experimental results in Sec XXX indicate that the “thinking” frequency of a commonly-used embodied AI system, OpenVLA\cite{kim24openvla}, is only 1-6Hz. Such low thinking frequency ($F$) further exacerbates the staleness problem, where actions generated at time $T + 1/F$ are based on input data from time $T$. Meanwhile, the GPU utilization is merely XXX\%.
\nonsense{To this end, many algorithm works~\cite{tinyvla, zhu2024llavaphiefficientmultimodalassistant, chu2024mobilevlmv2fasterstronger, zhao2025cobraextendingmambamultimodal} seek to use smaller models for inference but the throughput is still not ideal, and it is hard to fulfill the hardware's capability. Some humanoid robots (e.g., Figure AI's Helix~\cite{helix}) deploy another "fast thinking" system using an additional GPU to keep pace with the high-frequency robotic actions, without optimizing resource utilization.}

\nonsense{For the practical value of robotic agents, improving hardware efficiency is of equal importance to increasing the thinking frequency. However, the hardware remains underutilized currently, and effective strategies remain largely unexplored.}
% Except for the low thinking frequency, we observe that the current hardware is poorly used. 
The key obstacle to making good use of current hardware is that the \textbf{perception} and \textbf{generation} are predominantly implemented in a closed-loop pattern. 
%Specifically, each frame captured by the camera triggers a request within the embodied system. This request undergoes the thinking process to generate an action for the robotics. However, any additional requests generated during this period are discarded until the current request completes the overall action generation process.
The sequential computation of requests hinders possible throughput improvement. 
There is potential to improve the thinking throughput by increasing the parallelism, but two challenges appear. 
% First, by parallelizing the originally sequential requests, the latency of each request may be prolonged as depicted in~\autoref{fig:intro}(b), causing the output of the ``thinking'' process to be not timely, thereby decreasing the agent performance.
% Second, it is also non-trivial to provide high and stable output for agents because interferences across requests can be severe because all computations are executed on the same hardware device. Then the maximum throughput the hardware can provide is underestimated.
First, by parallelizing the original sequential requests, concurrent requests may interfere with each other using techniques such as CUDA multi-stream~\cite{cuda_stream}. This interference hinders the agent from providing stable and high throughput (e.g., the output interval has an 89.8\% fluctuation on average). Second, the latency of each request may be prolonged when paralleled as depicted in~\autoref{fig:intro}(b), causing the ``thinking'' process to compute on staled data, 
%and generate responses that are not timely, 
thereby decreasing the agent's accuracy.

% \idea{interference.}
%To achieve high throughput, we claim that it is beneficial to organize the computation of concurrent requests in a regular and controlled pattern (e.g., pipeline parallelism) to alleviate interference.
Our key insights are:  1) it is possible to alleviate interference while achieving high thinking throughput, 
by organizing concurrent requests in a regular and controlled pattern (e.g., pipeline parallelism).
%Meanwhile, %to maintain high thinking accuracy,
%To enable concurrent requests of agents to all compute based on the latest environment data and maintain high agent accuracy, 
%2) disaggregate the perception and generation steps. 
2) Disaggregating the perception and generation steps helps in maintaining the thinking accuracy. After the disaggregation,
%In this way, 
the generation phase can always be computed based on the fresh data exported from the latest perception phase, thereby maintaining high accuracy.

It is non-trivial to achieve the above design. 
While both perception and generation are intertwined on the same device, it is difficult to optimize the throughput through software pipelining. 
% Handling the complicated parallelism pattern of both perception and generation to achieve high throughput requires delicate design.
%The complicated parallelism of the two processes calls for delicate designs. %to maintain high throughput.
% It is helpful to achieve an efficient execution pattern by disaggregating perception and generation and managing their pipelines in a unified framework.
A promising strategy is disaggregating the two stages and managing their pipelines in a unified framework.
Moreover, resolving the data staleness problem requires effective algorithm adaptation in the disaggregation design. 
It is complex to instruct the data flow between perception and generation while maintaining accuracy. 
We therefore propose \sysname{}, an embodied AI system that enables high-frequency perception and generation without compromising agent accuracy. 
To increase parallelism while maintaining accuracy, we analyze the mainstream algorithms and disaggregate the perception and generation, while sharing a public context buffer between them. By interacting through the public buffer, concurrent generation phases benefit from the latest data.
To achieve high throughput while not harming agent accuracy, we propose an asynchronous pipeline executor to allow pipeline parallelism for both perception and generation. The executor targets to coordinate the execution pattern across different stages and we can then achieve a sweet point between high accuracy and high frequency.
% The platform focuses on decoupling and efficiently orchestrating perception and generation modules, tackling the computational bottlenecks inherent in sequential computation patterns. Given a generative AI algorithm provided by an algorithm developer, \sysname{} begins by identifying and analyzing the algorithm, separating it into distinct perception and generation components. Through proper splitting and pipelining of these components, \sysname{} enables deployment configurations that meet the accuracy and throughput requirements specified by users. 
% By balancing the trade-offs between decoupling and integration, \sysname{} ensures both high performance and robust reliability, making it a versatile and scalable solution for embodied AI systems operating in real-world scenarios.
This work makes three main contributions. 

\textbf{1) Detailed analysis of characteristics of prevalent embodied AI algorithms and systems.}
We analyze the inability to meet the high-frequency demands of real-world agents. 

\textbf{2) The co-design of algorithm and system for perception and generation disaggregation.}
The disaggregation design ensures that the output of agents is based on fresh input data.

\textbf{3) The design of asynchronous pipeline executor to achieve both high throughput and accuracy.} 
\sysname{} therefore provide controllable and stable output frequency.
% we validate the effectiveness of \sysname{} through extensive experiments, demonstrating its ability to improve throughput significantly without compromising performance, thus addressing key challenges in deploying embodied AI systems at scale. 
% Together, these contributions provide a foundational step toward bridging the gap between algorithmic development and system-level optimization in embodied AI.

% \paragraph{Diverse model architectures}
% \paragraph{Gap between slow brain and fast body}
% \paragraph{Inefficiencies in closed-loop control}

Experimental results %demonstrate the efficacy of \sysname{} in achieving these goals. 
show that \sysname{} improves the throughput of embodied AI systems with auto-regressive models by $3.05\times$ and diffusion-based models by $2.28\times$, while achieving almost the same agent accuracy compared to original sequential computation patterns ($102.7\%$ on average). 
These results highlight \sysname{}' capability to enhance system performance and scalability.
% , making it a versatile and robust solution for real-world embodied AI applications.
\section{Related Works}

{\bf Generative AI for robotics.}
Recently, more and more algorithm works focus on leveraging large generative models to implement embodied AI, covering scenarios including embodied question answering~\cite{EQA2018, Majumdar2024OpenEQAEQ, patel2024multillmqaembodiedexploration}, task planning~\cite{Huang2022LanguageMA, Huang2022InnerME, SocraticModels, EmbodiedGPT}, action planning~\cite{Jin2024ReasoningGV, Shen2023DistilledFF, SemGrasp, Describeexplainplanandselect, radosavovic2024humanoid, qian2025dispider} and so on. Within them, Google~\cite{rt1, rt2} demonstrates the potential of using an end-to-end vision-language-model to do semantic reasoning and generate actions in a discrete space. Following that, representative works such as OpenVLA~\cite{kim24openvla}, ICRT~\cite{yin2024context} and $\pi_0$~\cite{pi0} also employ LLMs to auto-regressively generate responses to interact with the environment. The agent interacts with the environment by predicting the next token.
Diffusion policy~\cite{chi2023diffusion} generates robot behavior by representing a robot’s visuomotor policy as a conditional denoising diffusion process and is also widely adopted in robotic agents~\cite{dasari2024ingredientsroboticdiffusiontransformers, DALL-E-Bot, Ma_2024_CVPR, xian2023chaineddiffuser} to generate actions. \nonsense{Some commercial-ready robots such as Helix~\cite{helix} and Nvidia's GR00T N1~\cite{gr00t} deploy two separate systems, namely `System 2' for slow-thinking and `System 1' for fast-thinking.}

{\bf System-level optimization.}
% \textcolor{red}{Requires refinement.}
There are plenty of works focusing on reducing the latency of executing generative models such as Large Language Models (LLMs). 
% For auto-regressive models, several strategies have been employed. 
To improve the effectiveness of handling concurrent LLM requests, scheduling techniques such as batching inference requests, as explored in prior works~\cite{agrawal2023sarathi, Orca2022, guldogan2024multibinbatchingincreasingllm}, have significantly increased the throughput of LLM serving systems.
For better resource utilization, different phases of the generation process are disaggregated and utilize separate resources. 
For example, some works~\cite{DistServe, Splitwise, Hu2024MemServeCC} disaggregate the prefill and decoding phases to eliminate the interferences between them. \nonsense{To mitigate the inference using multi-stream on a single device, task prioritization and CUDA stream priorities techniques are leveraged in some previous works~\cite{pang2023efficient, xiang2019pipelined}, and these techniques are effective for priority-sensitive tasks and requests.}
% The prefill phase generates the first token of the response from the user's prompt in one step. The decoding phase then generates subsequent tokens step by step.
% 1. scheduing: Several prior works~\cite{agrawal2023sarathi, Orca2022} have explored batching inference requests for higher throughput.
% 2. disaggregate
% 3. pipeline inference
% Orca~\cite{} leverages continuous batching to improve the overall throughput of streaming requests and xx optimize diffusion models through xxx.
% Quantization is a promising technique
In providing structured parallelism for large-scale LLM inference,
% Lastly, existing literature also explore various parallelism strategies in large-scale model inference. 
PipeLLM~\cite{PipeLLM} and PipeInfer~\cite{PipeInfer} propose pipeline parallelism in serving distributed LLM, while many frameworks~\cite{Kwon2023EfficientMM, zheng2024sglang} also provide data parallelism and tensor parallelism. 
There are also frameworks exploring various parallelism strategies in serving diffusion models~\cite{li2024distrifusion, fang2024xdit, wang2024pipefusion} for better throughput.

\section{Background and Motivation}

This section introduces the background of embodied AI algorithms, discusses the weaknesses of current systems, and presents the opportunities to improve thinking throughput. %based on which we find challenges and opportunities to improve it.

\begin{comment}
Embodied AI algorithms are often generative, requiring many iterations (steps of thinking) to output a response.

1. The sequential closed-loop execution pattern → uneven GPU utilization → low execution frequency

2. Increase parallel/pipeline degree can improve utilization, but for each observation/request, the generated response is not optimal.

3. We find that the volatile context can be updated for each response generation, then the precision is maintained.

4. We then strive to improve the throughput, and find pipelining is much better than pure parallel.
\end{comment}

\subsection{Embodied AI Compute Pattern}

\begin{figure}
    \centering
    \includegraphics[width=.95\linewidth]{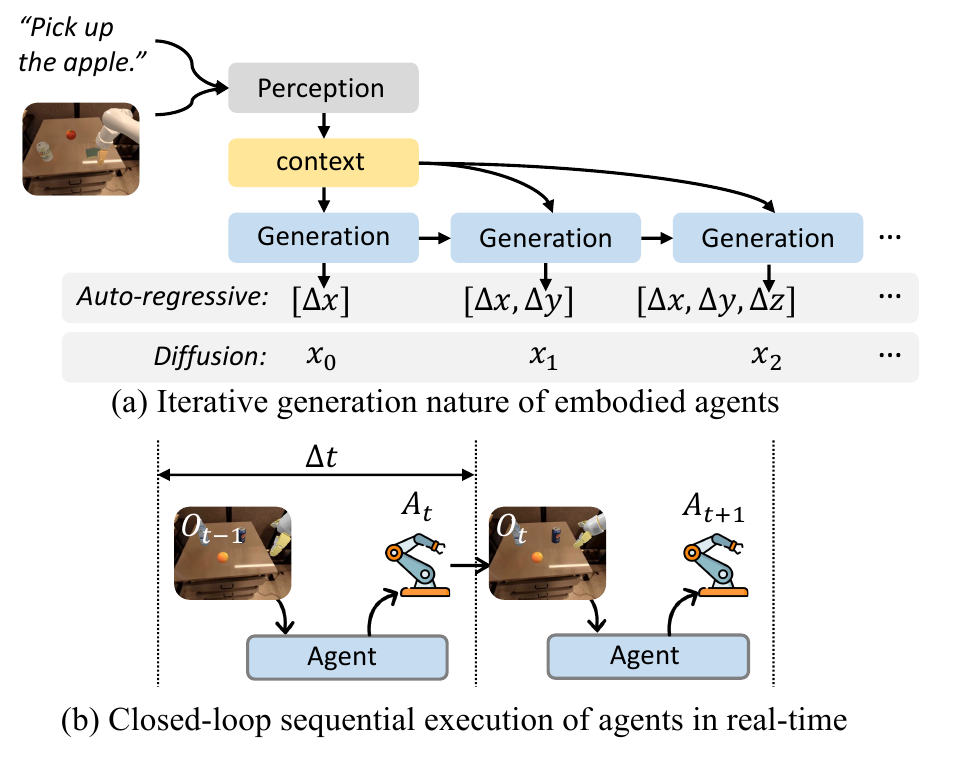}
    \vspace{-2mm}
    \caption{The compute pattern of embodied AI agents.}
    % \vspace{-2mm}
    \label{fig:iterative_gen}
\end{figure}

% \paragraph{Decompose the computation agent}
\autoref{fig:iterative_gen}(a) shows that the computation of an embodied AI agent includes two parts: \textbf{Perception} and \textbf{Generation}. The perception module receives multiple modal data as input and encodes them into a unified latent space, namely \textbf{context}. 
% The encoders are often modularized and pre-trained to obtain the representations of multi-modalities. 
% As for the generation module, there exist methods~\cite{} that use one-step encoders to generate the response. 
In the phase of generating actions or responses, current end-to-end embodied agents mainly leverage generative algorithms, because they have the ability to ``think'' iteratively~\cite{kim24openvla, rt2, chi2023diffusion} and perform better in reasoning and planning. 
There are typically two generative algorithms as agent policies. %as shown in~\autoref{fig:iterative_gen}(a):

1. Auto-regressive modeling: The generation module takes the latent context as input and generates responses in an auto-regressive manner. For robotic agents, each generated token represents a discrete value of the degree of freedom of the robot (e.g., the delta movement of the mechanical arm in the x-axis). For Google Robot~\cite{rt1, rt2}, the length of the generated response of each observation is fixed to be 7. Some mobile manipulation robots~\cite{walke2023bridgedata, jiang2023vima} have a degree of freedom of 5. There also exist other embodied agents that generate length-variable tokens~\cite{li2024llara, huang2024rekep}, acting as instructions to downstream action generation policies.

2. Diffusion-based modeling: Denoising Diffusion Probabilistic Models (DDPM) are a class of generative models where the output generation is modeled as a denoising process. The diffusion-based algorithms take Gaussian noisy data $x_0$ as input, while the context is also forwarded as the conditioning input. Through multiple iterations of denoising, a desired noise-free output $x_n$ is formed. The output can be predicted vectors of actions (e.g., the predicted robot actions~\cite{chi2023diffusion, octo} or trajectories~\cite{gu2022stochastic, Ma_2024_CVPR, xian2023chaineddiffuser}).

%\subsection{Poor Utilization of Closed-loop Computation}
\subsection{Poor Utilization with Current Pattern}
%When running in real-time, 
Current embodied AI agents cannot make good use of the equipped hardware resources.
%Run in a closed-loop and sequential manner, the agents continuously take the observation as input and forward the model to generate actions.
As shown in~\autoref{fig:iterative_gen}(b), the agent perceives and generates responses in a closed-loop sequential manner.
Define the interval between two adjacent actions generated $\Delta t$. Suppose the observation at time slot $t-1$ is $O_{t-1}$, then the corresponding response is $A_{t}=\texttt{Agent}(O_{t-1})$
% Typically, the agent receives observations of the environment as input and generate the corresponding response. 
This execution pattern ensures the bijection of observation and response. 
However, in real-time execution, this execution pattern cannot fully and evenly utilize the hardware's compute ability because different compute components of an agent have different resource requirements.
\begin{figure}
    \centering
    \includegraphics[width=.95\linewidth]{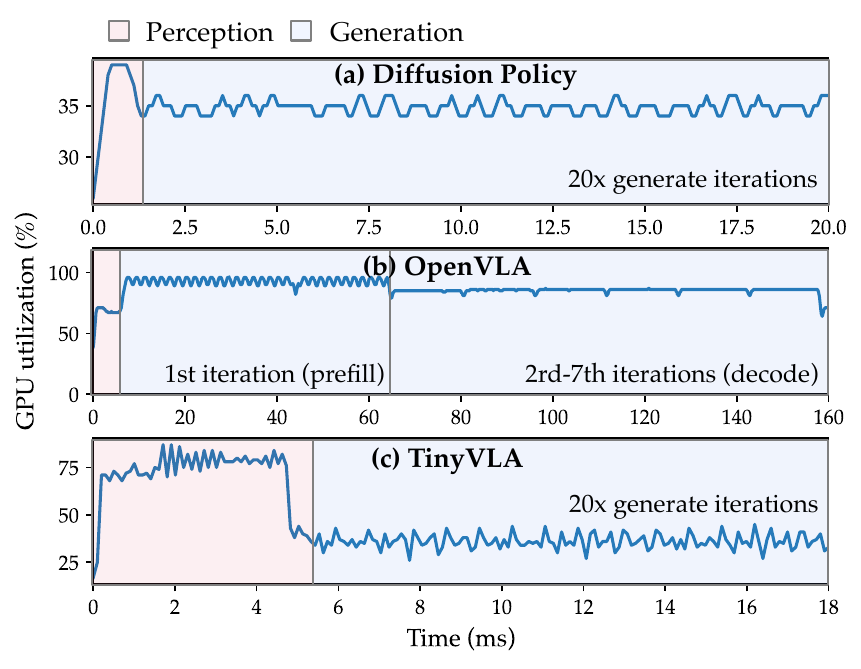}
    \vspace{-2mm}
    \caption{GPU utilization of robotic agents within a single request. The utilization is the SM Warp Occupancy profiled using Nsight system~\cite{nsight}. The profiled models are Diffusion Policy~\cite{chi2023diffusion}, OpenVLA~\cite{kim24openvla} and TinyVLA~\cite{tinyvla} respectively.}
    \label{fig:utilization}
\end{figure}

% \quan{We need to state the experimental setup.}

\autoref{fig:utilization} reveals the GPU utilization of typical robotic agents running on an NVIDIA RTX 4090 GPU within a time slot with the sequential execution pattern.
The average utilization of a diffusion-based agent (DP)~\cite{chi2023diffusion} is merely 34.9\%.
The utilization of OpenVLA~\cite{kim24openvla}, an auto-regressive agent is $79.2\%$, higher than Diffusion Policy. But we still observe the inefficiency within the decode iterations (lower than $75\%$). 
% Although the prefill phase has high GPU utilization, decoding is memory-bound and causes insufficient utilization of the hardware (\textcolor{red}{75\%} on average for the decode phase).
% \NX{confused}
We also observe the resource requirements of perception and generation are different. 
As shown in~\autoref{fig:utilization}(c), the utilization of the perception phase is 74.3\% and merely 36.3\% for the generation, revealing the low efficiency in the closed-loop sequential execution. Therefore, it is potential to increase the parallelism in computation to fulfill hardware utilization and increase agent throughput.
% However, it is challenging to manage concurrent requests to ensure high and stable throughput because requests are co-located on the same device and may experience severe interference.

\subsection{Challenges and Opportunities}
Although increasing the parallelism can potentially achieve a higher execution frequency or throughput, there are two challenges in maintaining high agent accuracy and exporting stable and equidistant action responses.

{\bf Challenge 1: Uncontrolled parallelism provides suboptimal throughput.}
When requests are simply computed concurrently (e.g., using CUDA multi-stream~\cite{cuda_stream}), the interference across requests is severe, leading to a suboptimal throughput. According to our experimental results, by increasing the degree of parallelism while not providing structured control, the throughput can be increased by $2.93\times$ on maximum compared with sequential execution. For comparison, when pipeline parallelism is applied, the throughput is increased by $3.47\times$. 
We thus demonstrate that \textbf{structured pipeline parallelism can provide high and stable throughput for embodied AI.}
It is challenging to manage the parallelism pattern for embodied AI systems because both perception and generation within the system are required to be parallelized for optimal execution. This leads to a complex parallelism configuration space, only by finding out a reasonable configuration within it can help to achieve both high throughput and agent accuracy.

\begin{figure}
    \centering
    \includegraphics[width=\linewidth]{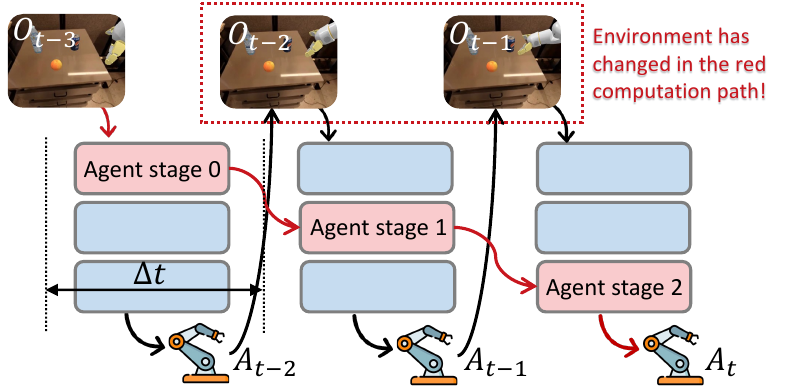}
    \vspace{-2mm}
    \caption{Concurrent execution of requests with pipeline degree = 3. The agent's computed output $A_t$ is based on the staled data $O_{t-3}$.}
    \label{fig:exec_pattern}
\end{figure}

\begin{figure}
    \centering
    \includegraphics[width=\linewidth]{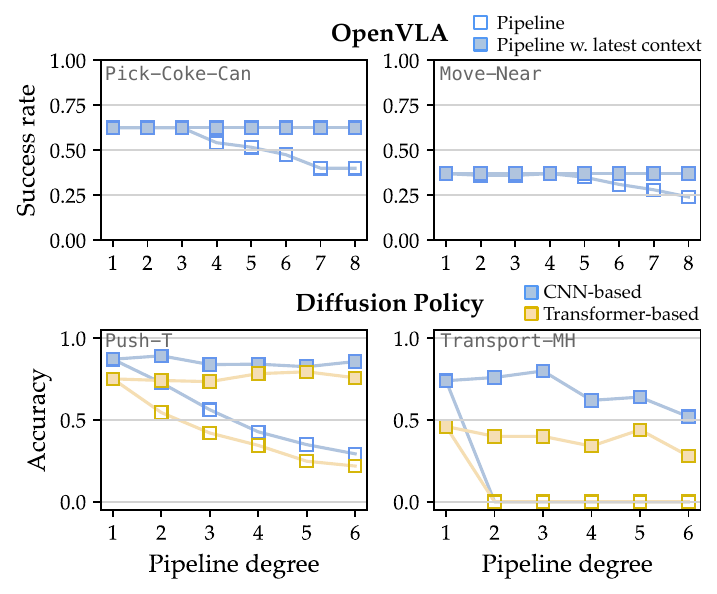}
    \vspace{-2mm}
    \caption{Agent accuracy with various degrees of parallelism. Pipeline degree = 1 refers to sequential execution. 
    % (a) and (b) are OpenVLA~\cite{} with \texttt{Pick-Coke-Can} task and \texttt{Move-Near} task, respectively. (c) and (d) are Diffusion Policy~\cite{} with \texttt{Push-T} and \texttt{Transport-MH}, respectively.
    }
    \label{fig:pipeline_degree}
\end{figure}

{\bf Challenge 2: Staled data degrades agent accuracy.}
% \idea{two challenges}
% This motivates us to increase the parallelism of agents to improve hardware utilization.
% We claim that it is the average low utilization hinders the improvement of execution frequency. 
% If the hardware utilization is high, we can then theoretically achieve a higher execution frequency or throughput.
% A naive idea is to increase the parallelism of processing requests to improve throughput. 
% Although increasing the parallelism can potentially achieve a higher execution frequency or throughput, we claim that the agent performance can be largely harmed because the latency of each request is prolonged and the generation is thus based on staled environment data, which can largely harm the precision.
With parallelism increased, the prolonged execution duration of each request can largely harm the precision.
% However, for each request, its latency is largely prolonged and the generated response may be stale with increased parallelism, thereby harming the precision of agents.
We illustrate this with an example depicted as ~\autoref{fig:exec_pattern}.
For example, when requests are executed concurrently with pipeline parallelism with a pipeline degree = 3 (i.e., the number of pipeline stages), three requests are computed simultaneously. With the interval of response generated still $\Delta t$, each response is computed as $A_{t}=\texttt{Agent}(O_{t-3})$, indicating that two latest observations $O_{t-2}$ and $O_{t-1}$ are skipped. Thus, although the increased parallelism may help to improve utilization, the data staleness may degrade the accuracy of agents.
The experimental results are shown in~\autoref{fig:pipeline_degree}. As we increase the pipeline degree, the accuracy of agents is degraded drastically. 
% \NX{can \sysname{} fix the staleness in the pipeline warmup stage? if not, figure 4 is a bad show case, \sysname{} can only remove the staleness in the steady stage of the pipeline right?}

% \paragraph{Algorithm with Context Update}
We identify that the key reason for the accuracy drop is that agents are generating responses according to stale data when executed in parallel. 
If all concurrent generations leverage the latest context, we observe that the agent accuracy is improved and comparable with the accuracy in sequential execution.
Therefore, \textbf{leveraging fresh data in the generation phase helps to maintain high accuracy.} This requires us to break the data dependency between perception and generation, so that the generation can be computed with the latest context.

\section{\sysname{} Methodology}
% 1. Algorithms with context update: Analyze the impact on algorithm precision (why we can do so)
% In this section, we first introduce how we decouple the perception and generation in embodied AI algorithms and share the public context with those components. Then we introduce

Based on the two opportunities above, we propose \sysname{}, an inference framework for embodied AI agents that boosts their execution while maintaining accuracy. 

\subsection{Overview of \sysname{}}

\begin{figure}
    \centering
    \includegraphics[width=\linewidth]{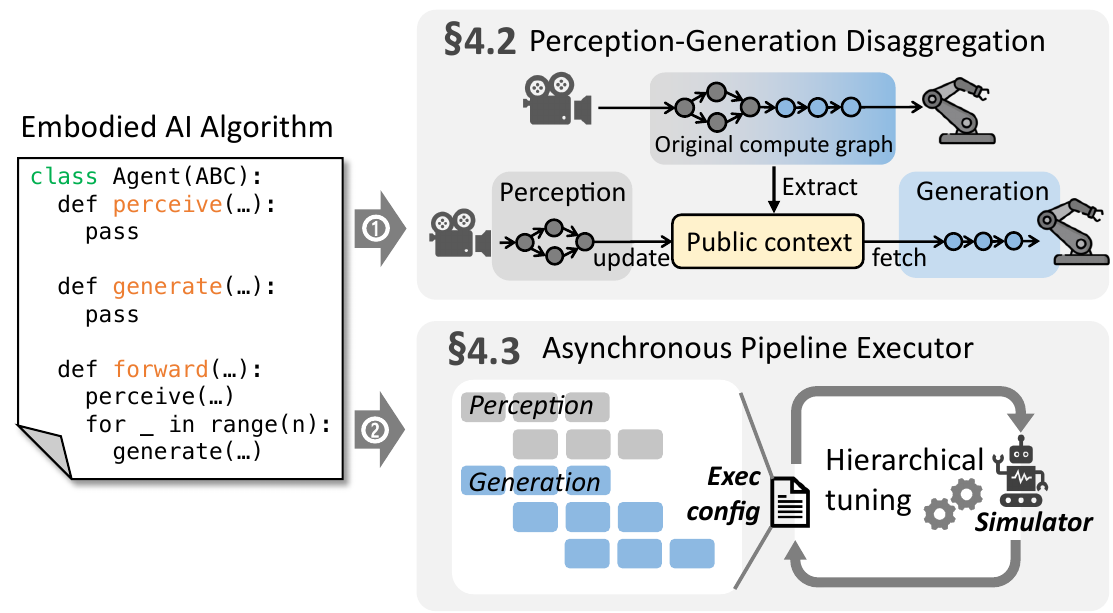}
    \caption{The systematic workflow of \sysname{}.}
    \label{fig:overview}
\end{figure}

\sysname{} leverages algorithm and system co-design to provide an efficient scheme for embodied agents.
\autoref{fig:overview} shows the systematic workflow of \sysname{}.

\sysname{} first modifies the compute pattern of embodied AI algorithms to achieve satisfying agent accuracy when the parallelism increases.
Given an embodied AI algorithm, \sysname{} takes a thorough analysis of its compute characteristics and disaggregates the perception and generation modules within.
A \textbf{public context} is extracted from the original compute graph to share between the two modules. 
In real time, the perception module updates the public context buffer and the generation module fetches from the buffer asynchronously.
This disaggregation design enables the generation module to always compute based on the latest environment context information.
Targeting the two mainstream categories of embodied AI algorithms: auto-regressive models and diffusion-based models, we explain how we identify the public context and maintain high accuracy in~\S\ref{sec:design1}.

% \sysname{} leverages optimized pipeline execution to increase the ``thinking'' frequency and adapts algorithms to achieve satisfying agent performance.

% Developers write templates for the perception module and the generation module, then annotate the volatile context.
% Based on the annotation, the \textit{Graph Reconstructor} separates the dependency of volatile context from the original compute graph.
% Instead of considering the computation of perception and generation as a whole, the \textit{Graph Reconstructor} decouples these two modules.

% The \textit{Pipeline Converter} and the \textit{Performance Filter}. 
% How do the perception module and the generation module interact with the public context buffer influences the execution pattern, thereby influencing the throughput and accuracy at the same time. 
The execution pattern of perception and generation influences the throughput and accuracy of the agent.
Compared with uncontrolled parallelism with no interference control on a single device, we claim that pipeline parallelism is more advanced in providing stable and predictable throughput and beneficial in maintaining accuracy.
Thus we propose the \textbf{asynchronous pipeline executor} (\S\ref{sec:design2}), which enables various algorithms to access the public context asynchronously, while in a controlled and defined manner. 
The execution behavior of the perception and generation pipelines is defined in the execution configuration, for which \sysname{} leverages a hierarchical tuner to achieve a sweet point between throughput and accuracy in the simulation environment.
% The \textit{Pipeline Converter} converts the original sequential computation graph into the pipelined form. The \textit{Performance Filter} profiles the configuration's accuracy on the simulation platform and the system-level gain (throughput/latency) on the target device. It filters out the configurations that do not meet the developers' requirements.

\subsection{Disaggregate Perception and Generation}
\label{sec:design1}
We first have a detailed analysis of the original sequential computation to disaggregate the perception and generation. Then we introduce how to form a \textbf{public context} to share between perception and generation.

\subsubsection{Compute Graph Analysis of Embodied Agents}

% Developers are responsible to separate the workflow of the perception module and the generation module, and annotate the volatile context data. 

An embodied AI algorithm consists of both perception and generation. For a request, the perception module takes multi-modal inputs from the environment as an observation and encodes them into a unified context using various models (e.g., image encoders~\cite{resnet, dosovitskiy2020image, chen2022pali} for vision inputs and natural language tokenizers~\cite{sennrich-etal-2016-neural, Japanese-and-Korean-voice-search, kudo-2018-subword} for text inputs). 
Although may comprised of multiple models, the computation of perception is finished in a one-step manner.

However, the generation phase is completed in an iterative manner. We analyze the original sequential compute graph of the generation phase for auto-regressive and diffusion algorithms, the two promising techniques in embodied AI.

\begin{figure}
    \centering
    \includegraphics[width=\linewidth]{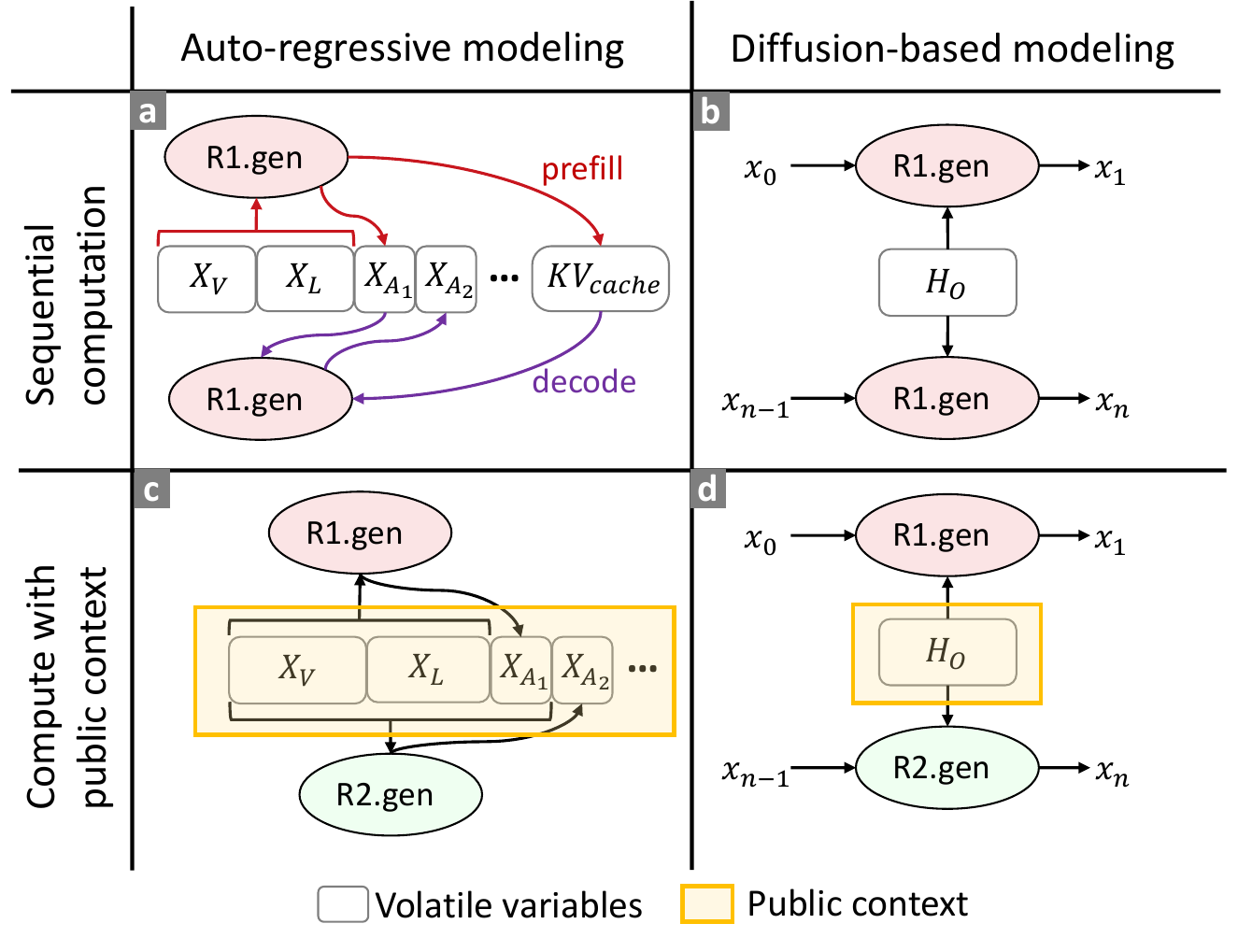}
    \vspace{-2mm}
    \caption{The compute graphs of the generation module. With sequential computation, the variables that request R1 use are consistent within the request, but are considered volatile when requests are concurrent ((a) and (b)). With the public context, generation steps from parallel requests (R1 and R2) can share the same latest public context for computation ((c) and (d)).}
    \label{fig:volatile_context}
\end{figure}

For auto-regressive models, the compute graph of a request R1's generation phase R1.gen is depicted in~\autoref{fig:volatile_context}(a), interacting with the token embedding  
% the context is first given by the perception module and represented as a sequence of token embeddings 
% $X=[x_{v_1}, x_{v_2}, \cdots, x_{l_1}, x_{l_2}, \cdots, x_{a_1}, x_{a_2}, \cdots]$, where $v,l,a$ represent the embedding is for vision observations
$X=[X_V, X_L, X_A]$, where $X_V, X_L, X_A$ represent the embeddings for vision observations, human language prompts and generated action tokens, respectively.
The embeddings $X_V$ and $X_L$ are the output of the perception module and remain unchanged within the request, while only the embeddings in $X_A$ are computed token-by-token auto-regressively. The consistency of $X_V$, $X_L$ and generated $X_A$ enables the original auto-regressive generation to be optimized to use KV-cache~\cite{Attentionisallyouneed, Kwon2023EfficientMM} to reduce computation. The generation is thus split into the prefill phase and multiple decode phases as Equation~\ref{eq:1}, where $l$ refers to the total length of $X_V$ and $X_L$, $l_a$ refers to the number of action tokens and $x$ refers to the embedding of a single token.
\begin{equation}
\label{eq:1}
\begin{aligned}
\colorbox{lightgray}{$x_{\leq l}$}&=\texttt{Perceive}(O)\\
\colorbox{lightgray}{$x_{l+1}, {KV_{cache}}_{\leq l}$}&=\texttt{Generate}(\colorbox{lightgray}{$x_{\leq l}$})\\
\colorbox{lightgray}{$x_{l+2}$}&=\texttt{Generate}(\colorbox{lightgray}{$x_{l+1}, {KV_{cache}}_{\leq l}$})\\
&\cdots \\
\colorbox{lightgray}{$x_{l+l_a}$}&=\texttt{Generate}(\colorbox{lightgray}{$x_{l+l_a-1}, {KV_{cache}}_{\leq l+l_a-2}$})
\end{aligned}
\end{equation}

% For each iteration in the auto-regressive model, the volatile context is the embedded latent, including the embedded of camera observations, human prompts, as well as the generated tokens for response.

\begin{figure*}
    \centering
    \includegraphics[width=\linewidth]{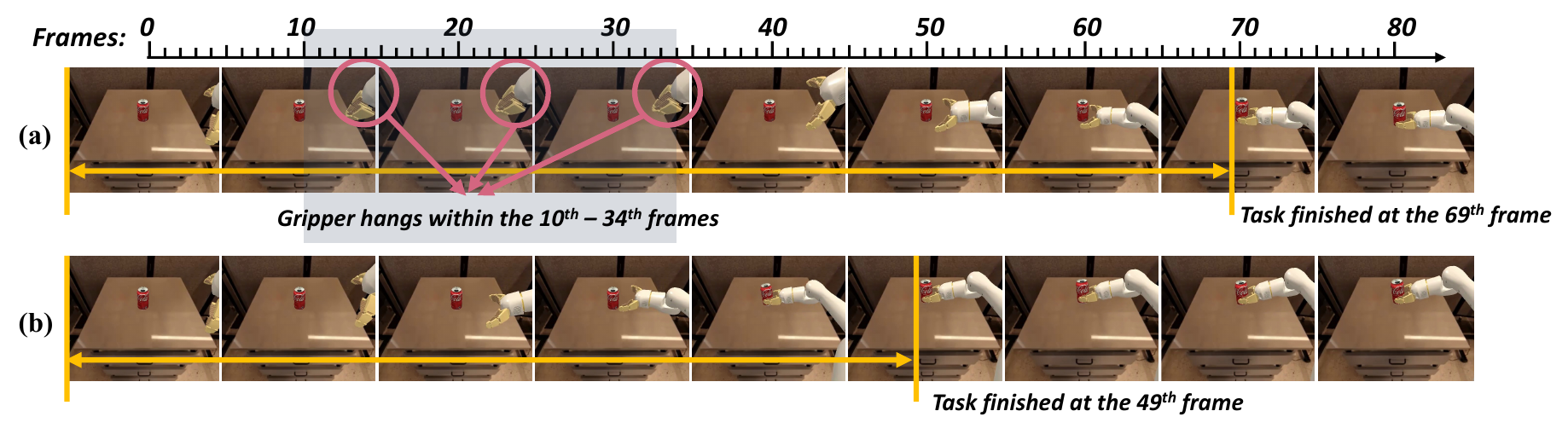}
    \vspace{-4mm}
    \caption{Behaviors of OpenVLA on the \texttt{Pick-Coke-Can} task. (a) The original sequential execution; (b) Execution with integrating the latest public context into the computation.}
    \label{fig:gripper}
\end{figure*}

The compute graph of diffusion-based models follows the denoising process.  
For diffusion-based models, the $H_o$ is consistent across different iterative steps as shown in~\autoref{fig:volatile_context}(b), which is the hidden intermediate of observations as shown in Equation~\ref{eq:2}. The $x_0$ is a random Gaussian noise and after $n$ steps of de-noising, the $x_n$ is the final output. 
\begin{equation}
\label{eq:2}
\begin{aligned}
\colorbox{lightgray}{$H_o$}&=\texttt{Perceive}(O)\\
x_1&=\texttt{Generate}(x_0, \colorbox{lightgray}{$H_o$})\\
&\cdots \\
x_n&=\texttt{Generate}(x_{n-1}, \colorbox{lightgray}{$H_o$})
\end{aligned}
\end{equation}

\autoref{fig:volatile_context} only shows the compute graphs of auto-regressive models and diffusion-based models. However, as long as the model is composed of perception and generation, we can extract the variables within which are shared by both perception and generation.

\subsubsection{Identify the Public Context}
As we increase the parallelism by breaking the compulsory closed-loop computation, the system can hold parallel requests (e.g., R1 and R2 in~\autoref{fig:volatile_context}(c) and (d)) to handle continuous observations instead of supporting only one request at a time.
The parallel requests can then see each other's contexts within their lifecycles. It is beneficial for each request to discard its private consistent context, and look for the most timely context to generate responses.
Therefore, we propose to establish a public context for parallel requests to use and fetch the latest environment information. It is thus crucial to identify the volatile context within each request that can be updated by the public one.

For auto-regressive models, the input in each generation step is annotated as volatile as colored gray in Equation~\ref{eq:1}, indicating that the KV-cache can not be shared across concurrent requests.
% , not only the variables in the context, but also the variables derived from it, which particularly pertains to the KVCache.
% Since the original consistent context becomes volatile within a request, at each iteration of generation, we use all the embeddings of tokens as $x_{t+i}=\texttt{Generate}(x_{\leq {t+i-1}})$, as shown in~\autoref{fig:volatile_context}(c). 
% \idea{Reverse the logic here.}
Thus, as shown in~\autoref{fig:volatile_context}(c), the whole context $[X_V, X_L, X_A]$ is regarded as the public context. 
In detail, to form the public context, the $X_V$ and $X_L$ are derived from the latest output of the perception module and the $X_A$ are updated by the concurrent requests: each request leverages a prefill from $[X_V, X_L, X_{A_{1\rightarrow i}}]$ to update the $X_{A_{i+1}}$.
The public context is therefore updated once new output from the perception module is calculated or new action tokens are generated.

For diffusion-based models, the $H_o$ is volatile as annotated in Equation~\ref{eq:2} and is considered as the public context.
Then different iterative steps in the generation phase of concurrent requests can all access the same public context.
The public context is derived from the perception module as the conditioning data for the generation policy network.

In summary, we regard the output of the perception module as the public context for diffusion-based models, and regard the output of perception module along with the generated action token embeddings as the public context for auto-regressive models. 
Due to the diversity of generative algorithms, we cannot simply consider the output of the perception module as the public context.

\subsubsection{Generation based on the Public Context}
% \textcolor{red}{Add a figure here.}
% Now we introduce the time information into computation
% In a specific frame, there exist multiple generation modules to be executed. 
With parallel requests sharing the public context to generate output action tokens, there is also a chance that parallel requests share the computation.
We explain how \sysname{} optimizes the generation by showcasing a scenario in which two parallel requests R1 and R2 are both in the generation phase of an auto-regressive model. 

Suppose the computation of the $i$-th iteration of R1 and the $j$-th iteration of R2 are to be executed at the same time, represented by $\texttt{Generate}(x_{\leq {l+i}})$ and $\texttt{Generate}(x_{\leq {l+j}})$, respectively. Then these computations can be merged into a single computation on the public context: 
the result of prefilling the first $l+i$ tokens can be extracted from prefilling the first $l+j$ tokens when $i$ < $j$. This holds true when the transformer applies causal masking, because causal masking ensures that the computation of the hidden state $h_{l+i}$ (for token $x_{l+i}$) depends only on tokens $x_1, x_2, \cdots, x_{l+i}$, and not on tokens $x_{l+i+1}, \cdots, x_{l+j}$.
Therefore, the hidden state $h_{l+i}$ is identical whether prefilling up to $l+i$ tokens or $l+j$ tokens.
% when $i$ < $j$, the result of $\texttt{Generate}(x_{\leq {l+i}})$ can be extracted from the result of $\texttt{Generate}(x_{\leq {l+j}})$.
% Since $x_{\leq {t+j}}$ and $x_{\leq {t+j}}$ are within the same public context, we seek to minimize the computation cost by removing redundant computation on the public context.
% We are glad to see that by the auto-regressive nature of the causal transformer, 

% \textcolor{red}{Check if necessary.} If the transformer has non-causal attention masking for tokens before $x_{l+i}$ and causal masking beyond $x_{l+i}$, the hidden states for the first $i$ tokens are still computed independently of tokens $x_{l+i+1}, \cdots, x_{l+j}$. Therefore, the result of prefilling the first $l+i$ tokens can still be extracted from prefilling the first $l+j$ tokens.

Thus, for auto-regressive models, when different iterative steps are overlapped and share the public context, only a merged unified computation as $\texttt{Generate}(X)$ is required to update the action token embeddings $X_A$. The merging of diverse iterative steps can significantly reduce computation for auto-regressive models. 
% For instance, if there are $N$ iterative generation steps from $N$ parallel requests to be overlapped, the merged execution can employ a single prefill to replace the $N$ prefills on the public context.
Since the denoised data is not identical across different requests in diffusion-based models, the computation merge is not applied to diffusion policies.

\subsubsection{Effectiveness of the Public Context}

\autoref{fig:gripper} reveals the real execution process w/wo our public context design, using OpenVLA~\cite{kim24openvla} on the Simpler simulation environment~\cite{li24simpler}, with the task of \texttt{Pick-Coke-Can}. The action window is 80 frames and the original sequential execution finishes the task at the 69th frame, while the algorithm with context update finishes the task at the 49th frame, indicating shorter steps to finish the task. We observe that the sequential computation pattern encounters an obstacle of the gripper hanging within the 10th frame to the 34th frame (the gripper keeps closed and the robotic arm keeps stale). 
This is because the original sequential computation does not incorporate the $X_A$ into computation and thus may hang when output actions are similar. 
With $X_A$ also considered as a part of the public buffer, all concurrent requests can see the latest generated tokens, which we claim is beneficial in dynamic environments.
% \idea{[Maybe there are some pronouns to express this case, that is the LLM outputs the same thing when inputs are similar.]}

Within all the tasks that are judged as successful, \sysname{} can reduce the number of frames by 20.5\% compared with sequential execution in the \texttt{Pick-Coke-Can} scenario. 
Note that we do not modify any model weights through fine-tuning.
% \textcolor{red}{No diffusion results.}
% , and thus the range of motion of the robotic arm within each frame is identical.

\subsection{Asynchronous Pipeline Executor}
\label{sec:design2}

% Since we have updated the computation graph through the graph reconstructor, we seek to increase the parallelism to parallelize the execution of the reconstructed graphs.
Since we have disaggregated the perception and generation, we must instruct the data flow between those two components and decide how the public context is shared between them to maintain accuracy. 
Besides, we also aim to provide a high and stable throughput.

The core concept of balancing throughput and accuracy is to establish deterministic \textbf{frames} and execute by frames. Therefore, the asynchronous pipeline executor enables the pipeline parallelism of both perception and generation and organizes different stages of perception and generation into a \textbf{frame}.
We first determine the interaction logic of perception and generation on the public context, by providing a public context buffer design. Then we employ a hierarchical tuning process that can automatically determine the optimal pipeline configuration for both high agent accuracy and throughput. 
% We leverage the asynchronous pipeline executor to pipeline parallel requests. It aims to achieve a sweet point between maintaining agent performance and achieving high throughput.
% The executor provides an interface for users to construct frames.

% The minimal unit in computation is \textbf{frame}.

\subsubsection{Interact with the Public Context Buffer}
\begin{figure}
    \centering
    \includegraphics[width=.98\linewidth]{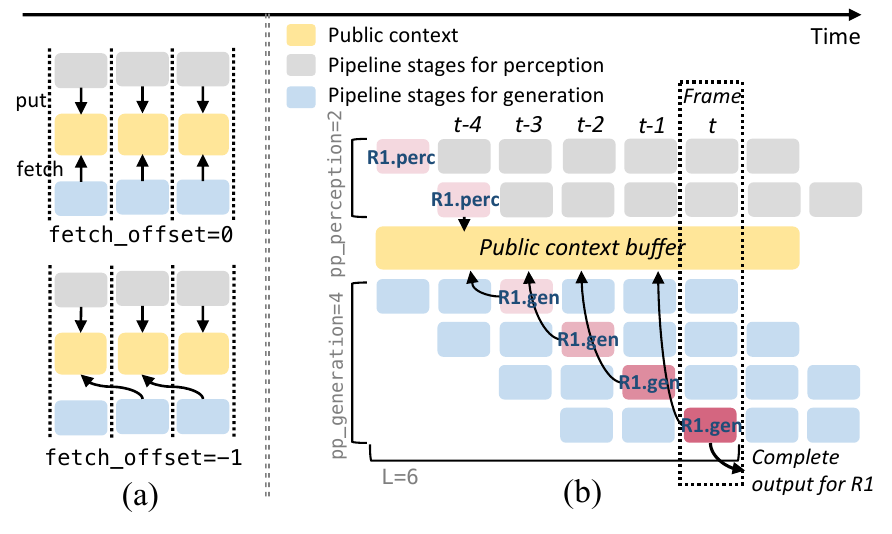}
    \caption{Illustration of the asynchronous pipeline executor. (a) Different patterns of accessing the public context buffer. (b) A configuration of pipelines for perception and generation. The complete compute process of the request R1 is marked red. In the generation phase of R1, the four generation stages compute on different public contexts (with different depths of color) because they are in different frames. The offset for generation to fetch the context buffer is -1 in this illustration.}
    \label{fig:pipeline}
\end{figure}

% \paragraph{Pipeline executor interface}
% We provide two interfaces to determine the compute pattern within each frame of the executor. First, we decide how perception and generation are pipelined. We provide the
% \texttt{perception.split(pp\_perception, args)} API to split a perception module into \texttt{pp\_perception} stages for pipelining. The counterpart for generation is \texttt{generation.split(pp\_generation, args)}. The \texttt{args} indicates the specific rule to split the model.
We first establish a public context buffer to store the public context specific to each frame. Then the executor decides how the public context is shared between perception and generation. 
As shown in~\autoref{fig:pipeline}(a), within each frame, the perception module puts the current computed context into the buffer to keep it always containing the latest environment information. Meanwhile, for the generation, it is allowed to fetch and compute on the public context of a past frame with an offset \texttt{fetch\_offset}.
This design helps to increase the diversity of parallelism between perception and generation and is effective for some auto-regressive models (\S\ref{sec:eva-pipe}). We apply the double-buffering design to avoid the locking and unlocking overhead on the public context buffer. Because the context's size is negligible compared with the total device memory, the overhead of this double-buffering design is trivial.

When \texttt{fetch\_offset} is 0, which means perception and generation are operating on the same public context within a frame, they have to be executed in sequence because of the data dependency. This ensures that within a frame, the generation receives the latest context information because no actions are taken to change the environment within the frame. 
When the \texttt{fetch\_offset} is set to -1, the generation within a frame fetches and computes on the public context that is put by the perception module in the last frame. In this scenario, the perception and generation can run in parallel.
A negative \texttt{fetch\_offset} indicates that perception and generation asynchronously access the public context buffer.
% \idea{delete: and the executor applies a double buffer design to achieve this.}
Different algorithms have different affinities to this offset parameter. 
% \textcolor{red}{Better explanations here?}
It is set to 0 for all diffusion-based policies and -1 for all auto-regressive policies.
% We can use the \texttt{put(frame\_id)} API to instruct which frame's public buffer it is going to update. The counterpart for generation is \texttt{fetch(public\_buffer.frame\_id)}. 
% \sysname{} then pipelines the perception and generation separately as depicted in~\autoref{fig:pipeline}.

% The asynchronous pipeline executor organizes the stages of both perception and generation into a frame. As depicted, any input observation has to experience at least \texttt{pp\_perception} frames to put results to the public context and a complete generation phase would experience \texttt{pp\_generation} frames.

% Three parameters that determine the final pipeline pattern and the computational logic.
% We only profile the triplets in the search space that satisfy the requirement of data staleness.

\subsubsection{Hierarchical Tuning for Pipeline Parallelism}
The complete configuration space for pipelining is large because we have not only the number of pipeline stages and the method to split the model to decide, but also need to consider both perception and generation. We therefore leverage a hierarchical tuning process to determine how these two modules are pipelined. We first determine the number of pipeline stages and then finetune the partitioning method, considering both perception and generation.

\paragraph{Determine the number of pipeline stages.} 
% \textcolor{red}{This caption is too plain.} 
We first determine the number of pipeline stages for perception and generation because they drastically influence the throughput as well as agent accuracy. Specifically, the \texttt{pp\_perception} is the pipeline degree of the perception module. Any input observation has to experience \texttt{pp\_perception} frames to put results to the public context buffer. 
The \texttt{pp\_generation} is the pipeline degree of the generation module. The \texttt{pp\_generation} is identical to the number of frames that a complete generation phase would experience.
Thus, within each frame, there are \texttt{pp\_perception} stages for perception and \texttt{pp\_generation} stages for generation from different requests, and these stages are computed concurrently. All generation stages can access the public context buffer and take the latest public context as input as shown in~\autoref{fig:pipeline}(b).

\sysname{} performs a grid search for these two parameters for pipelining to determine the optimal configuration that meets the requirements of users. 
% \idea{delete: The value of $\texttt{pp\_perception} + \texttt{pp\_generation}$ represents the number of frames between an agent's full response and the most distant input corresponding to it, which reflects the staleness of observation data for each generated response.} 
Define $L$ as the number of frames that a request needs to experience. Then the request R1 shown in ~\autoref{fig:pipeline}(b) experiences $L=\texttt{pp\_perception} + \texttt{pp\_generation}=6$ frames to complete output.
% \idea{Each request requires $\texttt{pp\_perception} + \texttt{pp\_generation}$ stages to generate actions. For example, the request R1 shown in ~\autoref{fig:pipeline}(b) needs 2 perception stages and 4 generations stages to complete output, and each of the generation stage fetches the the latest context from the public context buffer.}
We establish an upper bound for $L$ in the grid search to prevent searching on extremely poor configurations.

\paragraph{Fine-tune the partitioning of computations.}
% \textcolor{red}{Not good enough.}
The partitioning of the perception and generation module in the grid search follows the uniform distribution, where each computation stage of the perception phase consists of identical computation flops, and so does the generation phase. 
This ensures that the concurrent computation within a frame can be overlapped as much as possible, which maximizes the throughput.
Within the generation of each request, the iterative steps are uniformly distributed to the multiple stages. Suppose the number of iterations for generation of request R1 in~\autoref{fig:pipeline}(b) is 100, then every 25 iterations are computed on the public context of frame $t-4$, $t-3$, $t-2$ and $t-1$ respectively.
% To fully utilize the hardware within each frame, the concurrent computation should be overlapped as much as possible, which requires a reasonable model split mechanism. 
% It is non-trivial to split the perception module or the generation module properly because the computation within each module may be uneven. For example, within an iterative step of the generation, the compute intensity of different model layers varies; within the complete generation phase, splitting different numbers of iterations to different stages can also cause imbalanced workloads. 
We find that by adjusting the mapping of computation onto multiple stages, both the throughput and accuracy of agents are impacted. It is feasible to sacrifice a certain amount of throughput and obtain additional accuracy gains.
% Most importantly, the split method influences the performance of agents.

After the pipeline degree of perception and generation is determined, \sysname{} applies fine-tuning on the mapping of the computation to different pipeline stages. 
Specifically, the distribution is controlled by a skewness weight $\alpha$, and the amount of computation in the $i$-th stage is multiplied by the weight $e^{i\alpha}$ and then normalized with other stages. When $\alpha>0$, more computations are executed by the later stages, which endows the whole generation process to incorporate more fresh context data.
For example, when $\alpha=0.5$, $46$ iterative steps are computed on the public context within frame $t-1$, indicating leveraging more on the freshest data.

% \subsection{Performance Filter}
% Pipelining and compute pattern extraction: Performance alignment (tradeoff between efficiency and accuracy)

\section{Implementation}

We implement \sysname{} with 4,000+ LoC of Python. It receives the PyTorch models wrapped in \texttt{torch.Module} and automatically slice models into multiple stages. For transformer models that are not implemented in PyTorch, \sysname{} converts it into x-transformers~\cite{xtransformers}. 
On Nvidia GPUs, since the compute pattern of perception and generation in embodied AI is fixed, 
we capture kernel launches with CUDA graph~\cite{2018cudagraph} and all computations on GPU within a frame are launched as CUDA graphs. Each CUDA graph is bonded to a specific CUDA stream\cite{cuda_stream} to support concurrency within the frames.

{\bf Optimizations.}
\sysname{} applies multiple optimizations for the execution and resource management within frames.
% We then analyze the operations within a frame to perform further optimizations.
% This process is responsible for optimizing the computation within a ``thinking'' frame. 

1. Memory management: When the model size is too large for the GPU's device memory to hold, we apply the weight offloading feature to reside some model parameters on the CPU at runtime. \sysname{} leverages HuggingFace's accelerate~\cite{huggingface2023accelerate} library to automate the weight offloading procedure.

2. Batched execution: When multiple concurrent compute graphs are identical within a frame, we batch the input to allow the framework to use more efficient kernels for computation.

3. Graph launch sequence: Since a CUDA graph may be launched multiple times within a frame for the generation phase, in order to overlap the launch overhead of CUDA graphs on CPU, \sysname{} alternately launch CUDA graphs from different streams.

% \idea{[This is more like introducing a toolbox including many optimization methods]}
% 1) Memory management: Buffer reuse and weight offloading. KV cache (no need anymore)
% 2) Batched execution: When computations in frames Enabling batching while possible.
% 3) 
\section{Evaluation}

In this section, we evaluate \sysname{} in improving the throughput of embodied AI systems while maintaining accuracy.

\begin{table}
\footnotesize
\centering
\caption{\label{tab:system} Hardware and software specifications.}
\begin{tabular}{ll}
\hline
\multicolumn{2}{c}{System Overview}               \\ \hline
CPU              & AMD EPYC 7763 64-core \\
GPU              & Nvidia RTX3090  \& RTX4090         \\
Runtimes         & CUDA 12.4          \\
Libraries        & PyTorch 2.2.1                  \\ \hline
\end{tabular}
\end{table}

\begin{table}[]
\footnotesize
\centering
\caption{\label{tab:models} Agent models in evaluation. ($\star$ represents that the agent generates with the auto-regressive policy and $\circ$ represents the diffusion policy.)}
\vspace{2mm}
\resizebox{0.48\textwidth}{!}{%
\begin{tabular}{c|c|c|c}
\hline
Agent & Perception & Generation & Open-sourced \\ \hline
OpenVLA $\star$~\cite{kim24openvla} & \begin{tabular}[c]{@{}c@{}}DinoV2~\cite{oquab2023dinov2} + \\ SigLIP~\cite{zhai2023sigmoidlosslanguageimage}\end{tabular} & OpenVLA & Yes \\ \hline
RT2$^\ast$ $\star$~\cite{rt2} & ViT-large~\cite{chen2022pali} & LLama2~\cite{touvron2023llama2openfoundation} & \begin{tabular}[c]{@{}c@{}}Adapted \\ in LLaRa~\cite{li2024llara}\end{tabular} \\ \hline
DP $\circ$~\cite{chi2023diffusion} & ResNet18~\cite{resnet} & Transformer-S & Yes \\ \hline
DP-CNN $\circ$~\cite{chi2023diffusion} & ResNet18 & UNet-S~\cite{U-Net} & Yes \\ \hline
DP-plus $\circ$~\cite{chi2023diffusion} & 4$\times$ ResNet18 & Transformer-L & Yes \\ \hline
DP-CNN-plus $\circ$~\cite{chi2023diffusion} & 4$\times$ ResNet18 & UNet-L & Yes \\ \hline
TinyVLA $\circ$~\cite{tinyvla} & Pythia~\cite{biderman2023pythia} & Transformer & No \\ \hline
\end{tabular}
} % end of resizebox
\end{table}

\begin{figure*}
    \centering
    \includegraphics[width=\linewidth]{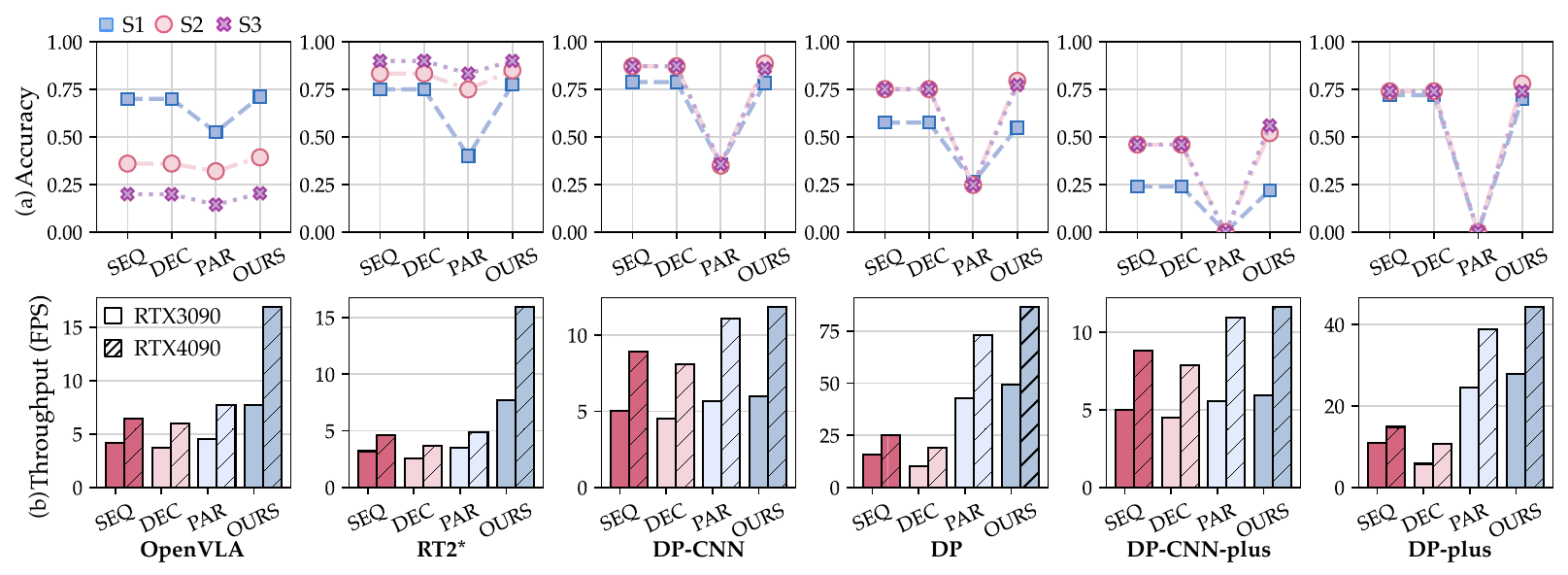}
    \vspace{-4mm}
    \caption{The accuracy and throughput of embodied robotic agents. \textbf{Accuracy}: The S1-S3 represents the average success rate in three scenarios: \texttt{Pick-Coke-Can}, \texttt{Move-Near} and \texttt{Open-Drawer} for OpenVLA, and  \texttt{Sweep-wo-Exceeding}, \texttt{Scene-Understanding} and \texttt{Visual-Manipulation} for RT2$^\ast$. For Diffusion Policy models, S1-S3 represents the result on models that perform $40, 100, 160$ diffusion steps on the \texttt{Push-T} task, respectively. \textbf{Throughput}: We anchor DP and DP-plus to 100 diffusion steps~\cite{chi2023diffusion} and anchor DP-CNN and DP-CNN-plus to 40 diffusion steps for faster inference.}
    \label{fig:overall_performance}
\end{figure*}

\subsection{Experimental Setup}

The system configuration is shown in~\autoref{tab:system}. To evaluate the effectiveness of \sysname{}, we use a suite of embodied AI agents that employ either auto-regressive models or diffusion models, as shown in~\autoref{tab:models}. This table also shows the specific perception and generation components for each model. 
% We tested 6 open-source models as our primary benchmarks, while the final mode, TinyVLA\cite{tinyvla}, is included as a case study to illustrate \sysname{}'s workflow in \autoref{section7.6}. 
% All models are quantized to the \texttt{bfloat16} datatype and executed on consumer-grade GPUs.

% \idea{[introduction of four agents?]}
\textbf{Agents and simulation environments.} We directly adopt the released parameters of the open-sourced models and run their trained tasks on their default simulation platforms. Specifically, OpenVLA is composed of two vision encoders (0.7B) and a 7B LLM backend and is tested on Simpler~\cite{li24simpler} with three scenarios,
%(\texttt{Pick-Coke-Can}, \texttt{Move-Near} and \texttt{Open-Drawer})
including 756 tasks in total. 
The original RT2~\cite{rt2} is a 560B model and is not released. We thus adopt RT2$^\star$, which is a released 7B model trained in LLaRa~\cite{li2024llara} in RT2's training style. It uses VIMA~\cite{jiang2023vima} and has 840 tasks. 
DP and DP-CNN are tested on the \texttt{Push-T} task, with 40MB and 505MB parameters separately. 
The plus versions of Diffusion Policy are tested on more complicated tasks in robomimic~\cite{robomimic2021} that two robotic arms and four cameras are adopted. The total model sizes are also expanded to 182MB and 527MB for DP-plus and DP-CNN-plus, separately.

\textbf{Compare targets.} We compare \sysname{} with three baseline execution modes: sequential, decoupled, and parallel execution modes. Sequential execution (SEQ) is the conventional pattern in existing embodied AI systems, where all incoming requests strictly follow the perception-generation workflow in a fully serial manner. With decoupled execution (DEC), perception and generation are fully separated into two different streams. During the generation phase, the system automatically fetches the latest perception results. Parallel execution (PAR) enhances system throughput by employing multiple workers, with each worker assigned an isolated thread and a CUDA stream. Each request is then assigned to an idle worker in real time.
%to generate actions in the same sequential manner, but in parallel across different workers. 
The request interval is aligned with the frame interval using \sysname{}, representing the same input frequency.

% While decoupled and parallel execution modes introduce only system-level modifications, \sysname{} incorporates both algorithmic and system-level designs. We use two metrics to evaluate the effectiveness of \sysname{} and three execution modes. The first metric, accuracy (i.e. the task success rate), assesses the quality of the generated actions. The second metric, throughput, measures the frequency at which actions are generated. 
% Both metrics are critical for embodied AI systems in real-world scenarios.  

\subsection{Overall Performance}

In this subsection, we evaluate the \textbf{accuracy} and \textbf{throughput} of the agents with different execution modes.

% In this subsection, we evaluate the \textbf{accuracy} and \textbf{throughput} of the agents under the four different execution modes: sequential, decoupled, parallel, and \sysname{}'s.
% We evaluate the \textbf{accuracy} and \textbf{throughput} of the models under the four different execution modes: sequential, decoupled, parallel, and \sysname{}'s. 
% \idea{delete?} The testing tasks include moving or picking up objects, opening drawers and sweeping. These activities exemplify common challenges that embodied AI systems encounter in realistic settings, thereby providing a relevant measure of performance under real environmental conditions. 

\paragraph{Accuracy.} Figure~\ref{fig:overall_performance}(a) shows the success rate of each model under different execution modes. The sequential mode naturally achieves high accuracy, as the perception step is always up-to-date before generation, so does decoupled execution mode. 
In contrast, the parallel mode suffers from a notable drop in success rate, on average by $80.22\%$, because individual requests take longer to process, causing each worker to rely on older (less fresh) perception results. 
Additionally, we observe that the accuracy drop problem is especially severe when the task is complicated. Compared with DP and DP-CNN, although the models are scaled up, DP-plus and DP-CNN-plus still suffer a complete disability (the accuracy drops to 0) in the parallel mode.

With \sysname{}, we resolve the data staleness problem by leveraging the public context design. This ensures that each generation phase benefits from the most recent perception results. As a result, \sysname{} maintains a success rate that closely matches that of sequential execution ($102.7\%$ on average across 6 different models).
%However, we also observe the phenomenon that when the number of inference steps is low, it is relatively hard for \sysname{} to maintain the same accuracy as the sequential mode. 
%For example, when the number of inference steps is 40, all diffusion policy models suffer a slight accuracy drop and the average accuracy drop is $1.82\%$. This is because that the number of effective diffusion steps is also insufficient when the total iterative steps are scarce. 
% \idea{explain so much?}

\paragraph{Throughput.} \autoref{fig:overall_performance}(b) presents the throughput of different models and execution modes when running on Nvidia RTX 4090 and RTX 3090 GPUs, respectively. \sysname{} increases the execution frequency by $2.29\times$, $3.01\times$ and $1.49\times$ compared with SEQ, DEC and PAR on average. Compared with the original 6Hz thinking frequency of OpenVLA, \sysname{} increases the frequency to 17Hz.

% \textbf{Decoupled Execution.} 
Compared to the sequential mode, the throughput under the decoupled execution mode decreases by $22.5\%$ on average.
% 7\%--27\% on RTX 4090 and 9\%--47\% on RTX 3090. 
This is because separating perception and generation into different streams can lead to interference, reducing throughput. 
% Since both steps utilize the GPU, running them concurrently without meticulous organization can lead to resource contention.
Additionally, in this execution mode, the perception component runs continuously, but the generation step does not consistently utilize its results. This leads to redundant computations and wastage of computational resources, further exacerbating the inefficiencies of this approach.

% \textbf{Parallel Execution.} 
Deploying multiple workers in parallel yields an increase in throughput of $1.06\times$--$2.93\times$, benefiting from the increased parallelism.
% for the RTX 4090 and $1.12\times$--$2.68\times$ for the RTX 3090. 
Diffusion-based models see the most pronounced gains (up to $2.93\times$). 
% This boost is primarily due to the asynchronous execution capabilities that parallel processing offers, allowing for more frequent updates and refinements to the diffusion steps without waiting for serial task completion. 
However, the overall throughput improvement is bounded by potential bottlenecks in GPU capacity and resource contention. Notably, for the two auto-regressive models, OpenVLA and RT2$^\ast$, the throughput gains are marginal, only achieving $1.06\times$--$1.29\times$.

% \textbf{\sysname{}.} 
By disaggregating perception and generation, \sysname{} outperforms the sequential approach across all tested models, boosting throughput by $1.32\times$--$3.48\times$ on RTX 4090 and $1.18\times$--$3.08\times$ on RTX 3090. 
The efficient pipelines adopted in the asynchronous pipeline executor of \sysname{} contribute to such throughput improvement.
Besides, the public context design enables auto-regressive models to share the generation across requests, further reducing the computation burden and improving the throughput. 
% \sysname{} adopts a structured and efficient pipeline. 
% Pipeline execution eliminates unnecessary runs of the perception modules that do not contribute to the generation tasks. 
% We provide a deeper analysis of these improvements in \autoref{sec:ex-par}.

\subsection{Effectiveness of Pipeline Parallelism}
\label{sec:ex-par}
\begin{figure}
    \centering
    \includegraphics[width=0.95\linewidth]{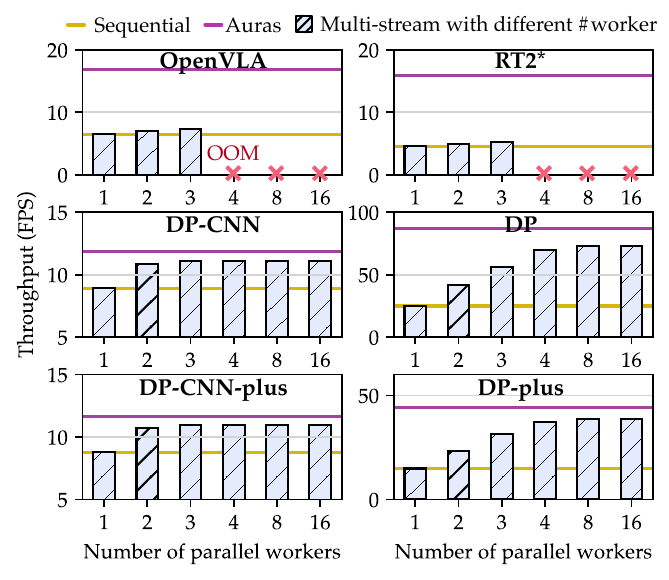}
    \vspace{-2mm}
    \caption{Throughput of agents under various parallelisms.}
    \label{fig:parallel_workernum}
\end{figure}

In this section, we conduct experiments to evaluate the effectiveness of pipeline parallelism in increasing the system throughput, compared with unstructured parallelism using multi-stream as shown in~\autoref{fig:parallel_workernum}.
% Simply increasing the number of concurrent workers does not conflict with the core concept of context update. 
% However, we demonstrate that the algorithm only works well together with the proper pipeline parallelism.
% To further investigate the impact of parallel execution mode on system throughput and highlight the effectiveness of \sysname{}'s design, 
% As shown in~\autoref{fig:parallel_workernum}, we evaluate the throughput of agents with different numbers of workers.

% \paragraph{Auto-regressive models.} 
We observe that although the throughput is increased with the number of worker streams in the parallel mode, there is still a gap compared with the pipeline parallelism achieved by \sysname{}'s asynchronous pipeline executor.
For the two auto-regressive models, OpenVLA and RT2$^\ast$, the throughput is only improved by a modest $1.11\times$ in the parallel mode, while \sysname{} improves the throughput by $2.20\times$--$3.29\times$.
% For the two auto-regressive models, as increasing the worker number from 1 to 3, the throughput of both OpenVLA and RT2$^\ast$ only improves by a modest $1.11\times$ in the parallel mode. 
% This limited gain underscores the heavy computational overhead associated with these models, while relying on a 7B large language model (LLM). Adding more workers merely amplifies GPU resource contention. 
% In contrast, as the purple line shown in \autoref{fig:parallel_workernum}, \sysname{} achieves $2.20\times$--$3.29\times$ higher throughput than the maximum throughput obtained by any parallel configuration.
For diffusion models, the throughput increases with the number of worker streams because of the comparably small model sizes. However, the gain diminishes as the parallelism degree reaches 16. \sysname{} easily surpasses the maximum achievable throughput by an additional $1.07\times$--$1.19\times$.
% The phenomenon holds for diffusion models: when the system transitions from one to two workers, the throughput improves by $1.22\times$--$1.67\times$, reflecting the effectiveness of parallelizing a lightweight model. However, as more workers are added, the throughput gain diminishes.
% \sysname{} surpasses the maximum parallel throughput by an additional $1.07\times$--$1.19\times$ for the tested diffusion models. 
We conclude that the benefits of pipeline parallelism over unstructured parallelism using multi-stream come from three aspects.

% \paragraph{Low interference between requests.}
\textbf{Low interference between requests. }
% Additionally, clear stage boundaries help minimize idle cycles: perception can gather fresh contextual information while generation is processing a previously calculated intermediate embedding. 
By separating perception and generation into well-organized stages, \sysname{} reduces interference between parallel requests. 
The interference can be evaluated by the job completion time (JCT) of each request. 
% When we impose the same input frequency of observations to parallel workers as \sysname{}, requests are queued and the JCT increases drastically in the parallel mode, indicating the simple parallelism of workers can not handle the stressful input.
When we impose the same input frequency of observations to parallel workers as \sysname{}, requests are queued and the JCT increases drastically.
If we impose just-in-fit input frequency, we observe that the JCT of each request is also longer than the JCT in the sequential mode.
For example, the JCT of OpenVLA is 153ms in the sequential mode, while in the parallel mode the JCT is 303ms when the number of workers is 2, which means the parallel two requests on the two workers are almost interwoven completely.
% \idea{We have to define the request elsewhere.}
% \idea{This data is not convincing.}

% \paragraph{Low memory consumption}
\textbf{Low memory consumption. }
We observe that for the two auto-regressive models, OpenVLA and RT2$^\ast$, the maximum number of workers we can deploy simultaneously is limited to 3 because of the limited VRAM on both the Nvidia RTX 4090 and RTX 3090 GPUs (24GiB). Although there is no need to replicate the model weights for multiple workers, each worker has to maintain its own CUDA graph for the agent model because the input and output buffers for different workers can not be shared. 
%Thus, the extra memory for CUDA graphs and corresponding buffers restricts the increase of parallelism. 
In contrast, \sysname{} only requires managing the CUDA graphs within a single frame and reusing them in different frames, thereby reducing the memory footprint. 
% The memory consumption does not grows with the pipeline degree. 
% \idea{kv cache?}

% \paragraph{Merged computation within frames}
\textbf{Merged computation within frames. }
% This boost primarily stems from two synergistic design elements in \sysname{}. 
% \idea{more benefits from pipeline here?} 
An additional optimization that helps \sysname{} to beat unstructured parallelism is the merging of computation based on the public context. It is applied in auto-regressive models. In transformer-based generative models that employ causal masking, the computation for the hidden state of the $i$-th token is independent of any tokens beyond $i$. This property allows a single large prefill to replace multiple smaller ones, substantially cutting down the number of repeated operations. 
% When combined, pipelining and shared context generation merging reinforce each other to alleviate GPU contention, eliminate redundant calculations, and significantly boost throughput in \sysname{}.

% \paragraph{Diffusion-based Policies.} For the diffusion policy models, we start with a single worker and scale up the number until we hit performance bottlenecks. 
% We test two diffusion models, each configurable with either one or two robotic arms under various parallel worker configurations. As shown in Figure \ref{fig:parallel_workernum}, when the system transitions from one to two workers, the throughput improves by $1.22\times$--$1.67\times$, reflecting the effectiveness of parallelizing a relatively lightweight model. However, as more workers are added, the throughput gain diminishes (i.e., exhibits a strong marginal effect), because additional concurrency induces increased competition for the same GPU resources.  \sysname{} surpasses the maximum parallel throughput by an additional $1.06\times$--$1.19\times$ for the tested diffusion models. This improvement comes mainly from \textbf{pipeline execution}.

\begin{figure}
    \centering
    \includegraphics[width=\linewidth]{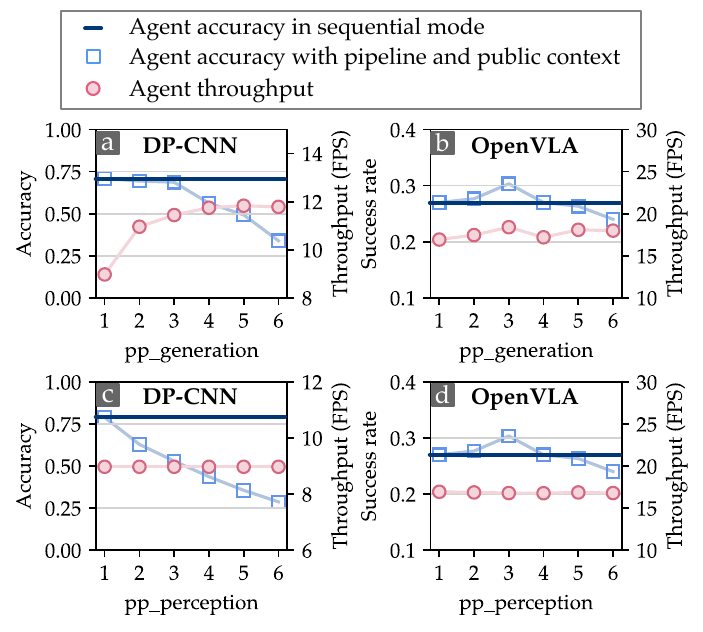}
    \vspace{-3mm}
    \caption{Throughput and accuracy tradeoff. The benchmark used in diffusion policy is \texttt{Push-T} and the benchmark used in the auto-regressive model is \texttt{Pick-Coke-Can}.}
    \label{fig:pp}
\end{figure}

\subsection{Effectiveness of Pipeline Tuning}
\label{sec:eva-pipe}
The asynchronous pipeline executor aims to find a sweet point between high throughput and accuracy. This subsection examines the effectiveness of the hierarchical tuning process in achieving a satisfying pipeline configuration.

\paragraph{Tuning the number of pipeline stages.}
\autoref{fig:pp} reveals the segments of the grid search when exploring the optimal number of pipeline stages. As shown in~\autoref{fig:pp}(a) and (b), we first adjust the \texttt{pp\_generation} and fix the $\texttt{pp\_perception}=1$. We observe that for diffusion policy, the agent accuracy decreases but the throughput increases with $\texttt{pp\_generation}$. 
% Thus, if the user can endure a $2\%$ accuracy drop, \sysname{} can select $\texttt{pp\_generation}=2$ for a better throughput. 
When \texttt{pp\_generation} is fixed to 1, increasing $\texttt{pp\_perception}$ causes a severe accuracy drop for diffusion policy as shown in~\autoref{fig:pp}(c).
% \NX{wrong placement, more sounds like the evaluation part?}

For auto-regressive models, the throughput is sensitive to neither $\texttt{pp\_generation}$ nor $\texttt{pp\_perception}$ alone, but the accuracy is affected by the summation $L=\texttt{pp\_perception}+\texttt{pp\_generation}$.
% \idea{[Rewrite.]} 
As shown in~\autoref{fig:pp}(b) and (d), when $L=3$, OpenVLA has a noticeable accuracy improvement. The non-monotonic relationship between pipeline parameters and accuracy holds for other auto-regressive models in our experiment, necessitating the complete grid search to find the optimal numbers of pipeline stages.

\begin{figure}
    \centering
    \includegraphics[width=\linewidth]{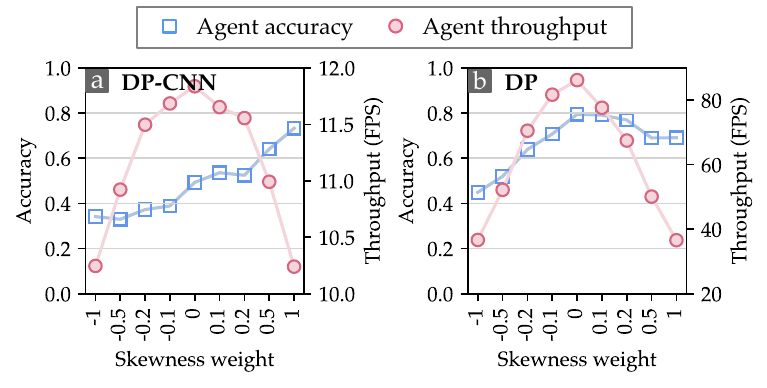}
    \vspace{-4mm}
    \caption{Accuracy and throughput with different partitioning skewness weight for generation ($\texttt{pp\_generation}=5$) on the \texttt{Push-T} task.}
    \label{fig:skewness}
\end{figure}

\paragraph{Fine-tuning the skewness of partition distribution.}
\autoref{fig:skewness} shows the agent accuracy and throughput of diffusion policy with different skewness when partitioning the generation phase. When the distribution is uniform, the overall throughput is the highest. However, with the increase of the skewness parameter $\alpha$, we notice an increasing trend in accuracy. This is because to generate a complete response, 64 of the 100 diffusion steps are generated according to the latest public context when $\alpha=1$, while the partitioning using uniform distribution only endows 20/100 diffusion steps computed on the latest data. 

The effectiveness of fine-tuning the partitioning is non-negligible. As shown in~\autoref{fig:skewness}(a), compared with the uniform distribution partitioning that has a $11.84$ FPS, a proper distribution ($\alpha=1$) can also maintain a $10.25$ FPS throughput and improves the accuracy by $23.98\%$.
As for the auto-regressive models, if the computation of generation is merged within each frame as introduced in~\S\ref{sec:design1}, adjusting the skewness of partitioning would not take effect.

\subsection{Scaling the Model}

% To work in diverse and complicated scenarios, embodied AI algorithms have to be adaptive to different inputs and outputs. The model structure provided by algorithm developers may vary drastically. For example, the input may vary: the number of input cameras, the number of history images as input and the resolution of observations; the output may vary: the number of robotic arms to control and the dimension of action. In this section, we use experimental results to show the performance of \sysname{} with different model structures.

To work in diverse and complicated scenarios, embodied AI algorithms have to be adaptive to different inputs, outputs, and model structures. 
The model provided by algorithm developers may vary drastically. 
For example, model structures can vary in parameter sizes and iterative steps.
% For example, the input may vary: the number of input cameras, the number of history images as input and the resolution of observations; the model structure may vary: the parameter sizes of perception and generation modules, the number of iterative steps; the output may vary: the number of robotic arms to control and the dimension of action. \textcolor{red}{Too long-winded}
In this subsection, we use experimental results to show the performance of \sysname{} scaling to different scenarios.

\begin{figure}
    \centering
    \includegraphics[width=\linewidth]{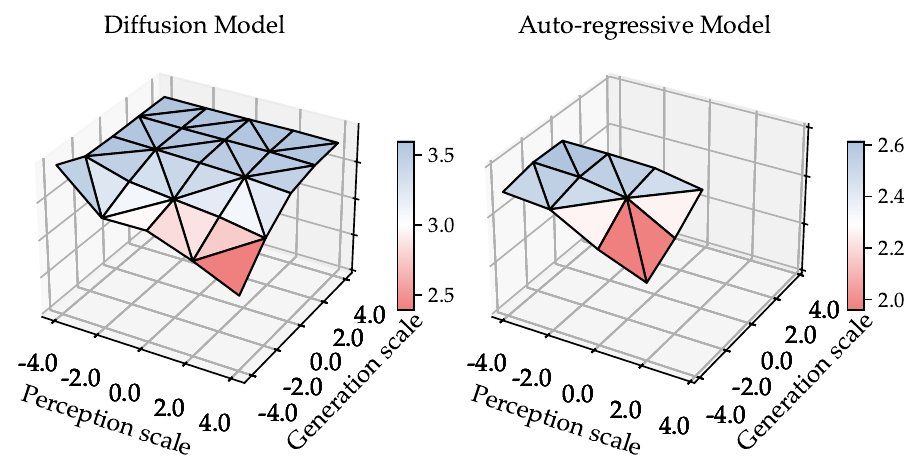}
    \vspace{-2mm}
    \caption{Speedup of \sysname{} compared with the sequential execution with different model parameter size. }
    \label{fig:scaling_model}
\end{figure}

% \paragraph{Increasing model size}
\paragraph{Various parameter sizes.}

% The distribution of parameters between perception and generation can vary greatly across different models. And within the same model, the parameters change when adapting to diverse inputs and outputs. 
We scale the parameters of two representative models: the auto-regressive OpenVLA and the diffusion-based DP-plus. We measure the throughput speedup of \sysname{} relative to the sequential baseline, as shown in Figure~\ref{fig:scaling_model}. 
We regard the original model size as 1, then apply scale factors of \{-4, -2, 2, 4\} to the perception and generation modules for exponential scaling, respectively. For instance, a scale factor of -2 for perception means its parameter size is $2^{-2}$ times the original size. 

Specifically, for OpenVLA, the speedup ranges from $1.53\times$ to $2.63\times$. For DP-plus, the speedup ranges from $1.93\times$ to $3.93\times$. 
Notably, both models share a similar trend: 
The speedup is proportional to the generation module's size and inversely proportional to the perception module's size.
% as the proportion of perception parameters increases relative to generation parameters, overall speedup declines. 
% This effect is evident in the lower-right region of the scatter plots, where larger parameter imbalance correlates with diminished parallel efficiency. 
This is because that perception module is less likely to be pipelined because it is typically finished in a one-step manner. 
%We find that increasing the perception model size often does not imply the increase of \texttt{pp\_perception}, otherwise, the accuracy drop is unavoidable. In this scenario, a larger perception model indicates more computation is executed sequentially, then the speedup becomes not prominent.
However, the generation module is executed iteratively naturally. Thus, increasing the model size of the generation module increases the possibility of pipelining. 
%With the \texttt{pp\_generation} increased, the speedup is also increased correspondingly.

\begin{figure}
    \centering
    \includegraphics[width=0.9\linewidth]{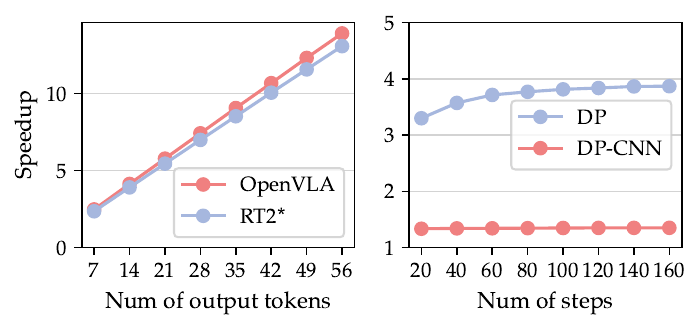}
    \vspace{-2mm}
    \caption{Speedup of \sysname{} with different numbers of iteration steps in the generation phase.}
    \vspace{-2mm}
    \label{fig:iteration}
\end{figure}

\paragraph{Various iterative steps}
The number of iterative steps, determined by the algorithm, directly influences the agent's accuracy. For example, more output tokens for an autoregressive model can be used to control more robotic arms, more degrees of freedom, or be applied in action ensembling~\cite{zhao2023learningfinegrainedbimanualmanipulation}. More iteration steps in diffusion indicate more fine-grained generation and higher accuracy. We evaluate the throughput speedup of \sysname{} with various iterative steps as shown in~\autoref{fig:iteration}.

For auto-regressive models, the speedup of \sysname{} is proportional to the length of output tokens. The core reason is that the throughput of sequential execution is inversely proportional to the number of iterative (decode) steps, while \sysname{} maintains a stable throughput because all decodings are integrated into one large prefill within a frame. The duration increase of a prefill with tens of token length growth is negligible when the input length is large, which is the case of large vision-language models. For instance, the length of embedded tokens encoded by a \texttt{ViT-large} of a 336px $\times$ 336px image is 576, which is $10\times$ larger than the output token length.

For diffusion models, we observe a minor speedup degradation when the number of iterative steps is small for the DP model. This is because when generation shrinks, the perception ratio expands, thereby reducing the opportunity for efficient overlapping across iterative generations.

\subsection{A Case Study in \sysname{}}
\label{section7.6}

The open-sourced models listed in~\autoref{tab:models} regard pre-trained vision encoders (e.g., ResNets~\cite{resnet} and vision transformers~\cite{dosovitskiy2020image}) to output intermediate embeddings. We also see a bunch of works integrating MLLMs to encode the images and prompts as the perception module~\cite{tinyvla, octo}. The action policies in these algorithms are also diffusion. 
The characteristic of the LLM-as-encoder model is that the computation within the perception module is comparably large.

\begin{figure}
    \centering
    \includegraphics[width=0.63\linewidth]{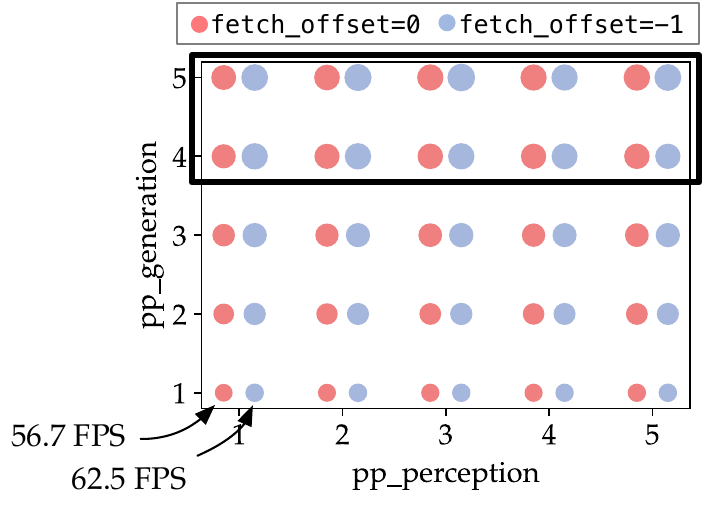}
    \vspace{-2mm}
    \caption{Throughput of \sysname{} on TinyVLA with different configurations.}
    \vspace{-2mm}
    \label{fig:tinyvla}
\end{figure}

We showcase the profiling result of TinyVLA~\cite{tinyvla} to exhibit how \sysname{} finds configurations that meet the throughput requirement of users. 
As shown in~\autoref{fig:tinyvla}, with different offsets for the generation to fetch from the public context buffer, the difference in throughput is not prominent. Then \sysname{} applies the grid search to find configurations that meet the throughput and accuracy requirement of users. We assume a 2$\times$ higher throughput is required and the \texttt{fetch\_offset} is set to -1, then the boxed configurations would be forwarded to the simulation platform to further test the accuracy.
% It first filter out the configurations that do not meet the throughput requirement before profiling the agent performance because evaluation in simulation environment is more time consuming. 
% As shown in~\autoref{}, if a user requires 2$\times$ higher throughput, only the boxed configurations would be forwarded to the simulation platform to test the accuracy.

% We also examine the GPU utilization after using \sysname{}. With the GPU utilization only xx on average in the sequential configuration ($[1, 1, 1]$), the average utilization comes to xx with a $[x,x,x]$ configuration and xx with a $[x,x,x]$ configuration.

\subsection{Startup Time}
The overhead introduced by \sysname{} is predominantly confined to the startup phase of the system. For auto-regressive models, 
\sysname{} initiates by performing a serial execution to generate the initial public context. Specifically, the startup time is 154ms for OpenVLA and 218ms for RT2$^\ast$. For diffusion models, the overhead is associated with the pipeline fill time. which are 40 ms and 67 ms for the DP and DP-plus models, respectively. 
The millisecond level of startup time is negligible in operating a real-world task.
% Meanwhile, for the DP-CNN and DP-CNN-plus models, the pipeline fill times are 279 ms and 282 ms, respectively.
\section{Conclusion}

In this paper, we addressed the limitations of traditional sequential computation patterns in embodied AI systems, particularly their inability to achieve the high "thinking" frequency required for real-world applications. To overcome these challenges, we introduced \sysname{}, an algorithm-system co-designed inference framework to optimize the inference frequency of embodied AI systems. \sysname{} disaggregates the perception and generation, and provides controlled pipeline parallelism for them. 
% Through efficient splitting, pipelining, and system-level optimizations, 
% \sysname{} boosts the "thinking" frequency of embodied AI systems while maintaining accuracy.
Experimental results demonstrated that \sysname{} improves throughput by 2.54$\times$ on average while achieving 102.7\% of the original accuracy, demonstrating its effectiveness and scalability.

\bibliographystyle{ACM-Reference-Format}
\bibliography{sample-base}

\end{document}